%% file: kdd2016.tex
\documentclass{sig-alternate} 
\pdfoutput=1 

\usepackage[T1]{fontenc}
\usepackage[utf8]{inputenc}
\usepackage{lmodern}

\usepackage{graphicx} 
\usepackage{color}
\usepackage{subcaption}
\usepackage[numbers]{natbib}
\renewcommand{\cite}[1]{\citep{#1}}

\usepackage{algorithm}
\usepackage{algorithmic}

\usepackage[english]{babel}
\usepackage{booktabs}
\usepackage{xspace}
\usepackage{complexity} 
\usepackage{verbatim} 
\usepackage{wrapfig} 

\newcommand{\pdftitle}{A Subsequence Interleaving Model \\ for Sequential Pattern Mining}

\usepackage[hidelinks]{hyperref}
\definecolor{MidnightBlue}{rgb}{0,0,0.4375} 
\hypersetup{
  pdftitle={\pdftitle},
  pdfauthor={},
  plainpages=false,
  colorlinks=false,
  linkcolor=black,
  citecolor=MidnightBlue,
  urlcolor=MidnightBlue,
}
\addto\extrasenglish{}
\addto\extrasenglish{}
\addto\extrasenglish{}

\numberwithin{equation}{section} 
\allowdisplaybreaks[3] 
\usepackage{enumitem}
\setitemize{noitemsep,topsep=0pt,parsep=0pt,partopsep=0pt,leftmargin=*}
\setenumerate{noitemsep,topsep=0pt,parsep=0pt,partopsep=0pt,leftmargin=*}

\newcommand{\cf}{\hbox{{cf.}}\xspace}
\newcommand{\etal}{\hbox{{et al.}}\xspace}
\newcommand{\eg}{\hbox{{e.g.}}\xspace}
\newcommand{\ie}{\hbox{{i.e.}}\xspace, } 
\newcommand{\st}{\hbox{{s.t.}}\xspace}

\newcommand{\etc}{\hbox{{etc.}}\xspace}

\newcommand{\abs}[1]{\lvert #1 \rvert}
\newcommand{\norm}[1]{\lVert #1 \rVert}
\newcommand{\deq}{\mathrel{\mathop:}=}

\renewcommand{\S}{§} 
\newcommand{\calI}[0]{\mathcal{I}}
\newcommand{\zB}[0]{\mathbf{z}}
\newcommand{\PiB}[0]{\boldsymbol{\Pi}}

\makeatletter
\def\@copyrightspace{\relax}
\makeatother

\newlength{\emstr}
\newcommand{\boldpara}[1]{%
\smallskip%
\par\noindent\textbf{\textit{#1}}\hspace{\emstr}
}%

\newdef{definition}{Definition}

\begin{document} 
%

\title{\pdftitle}

\numberofauthors{2}

\author{%
\alignauthor Jaroslav Fowkes
\and%
\alignauthor Charles Sutton
\and%
\end{tabular}\newline\begin{tabular}{c}
\affaddr{School of Informatics} \\
\affaddr{University of Edinburgh, Edinburgh, EH8 9AB, UK} \\
\affaddr{\{jfowkes, csutton\}@ed.ac.uk}
}




\def\@maketitle{\newpage
 \null
 \setbox\@acmtitlebox\vbox{%
\baselineskip 20pt
\vskip 2em                   
   \begin{center}
    {\ttlfnt \@title\par}       
   \end{center}}
 \dimen0=\ht\@acmtitlebox
 \unvbox\@acmtitlebox
 \ifdim\dimen0<0.0pt\relax\vskip-\dimen0\fi}

\maketitle
\begin{abstract} 
\input{./text/abstract.tex}

\end{abstract} 

\input{./text/introduction.tex}

\input{./text/related_work.tex}

\newpage
\input{./text/problem_formulation.tex}

\input{./text/experiments.tex}

\input{./text/conclusions.tex}

\section*{Acknowledgments} 
This work was supported by the Engineering and Physical Sciences Research 
Council (grant number EP/K024043/1).
{\small
\renewcommand*{\bibfont}{\raggedright}
\renewcommand{\bibsep}{0.6pt}
\bibliography{kdd2016}
\bibliographystyle{abbrvnat}
}
\balancecolumns

\end{document}

%% file: text/abstract.tex
Recent sequential pattern mining methods have used the minimum description length (MDL) principle to define an encoding scheme which describes an algorithm for mining the most compressing patterns in a database.
We present a novel subsequence interleaving model based
on a probabilistic model of the sequence database,
which allows us to search for the most compressing set of patterns without designing a specific encoding scheme. Our proposed algorithm
is able to efficiently mine
the most relevant sequential patterns and rank them using 
an associated measure of interestingness. The efficient inference in our model is a direct result of our use of a structural expectation-maximization framework, in which the expectation-step takes the form of a submodular optimization problem subject to a coverage constraint. We show on both synthetic and real world datasets that our model mines a set of sequential patterns with low spuriousness and redundancy, high interpretability and usefulness in real-world applications. Furthermore, we demonstrate that 
the quality of the patterns from our approach is comparable to, if not better than, existing state of the art sequential pattern mining algorithms.

%% file: text/introduction.tex
\section{Introduction}\label{sec:intro}

Sequential data pose a challenge to
exploratory data analysis, as large data sets of sequences are difficult
to visualise. In applications such as
healthcare (patterns in patient paths \cite{jay2004sequential}), click streams (web usage mining \cite{mobasher2002using}), bioinformatics (predicting protein sequence function \cite{wang2008sequential}) and source code (API call patterns \cite{zhong2009mapo}),
a common approach has been \emph{sequential pattern mining}, to identify
a set of patterns that commonly occur as subsequences of the sequences in the data.

A natural family of approaches for sequential pattern mining is to mine frequent subsequences 
 \cite{agrawal1995mining}
or closed frequent subsequences \cite{wang2004bide}, but these suffer from the well-known problem of pattern 
explosion, that is, the list of frequent subsequences is typically long, highly redundant, and difficult to understand.
Recently, researchers have introduced methods to prevent the problem of pattern explosion
based on the  \emph{minimum description length} (MDL) principle \cite{lam2014mining,tatti2012long}. 
These methods define an encoding scheme which describes an algorithm for compressing a sequence database
based on a library of subsequence patterns, and then search for a set of patterns that lead to the best compression of the database.
These MDL methods provide a  theoretically principled approach that results in better patterns
than frequent subsequence mining, but their performance relies on designing a coding scheme.

In this paper, we introduce an alternate probabilistic perspective
on subsequence mining, in which we develop a generative model of the 
database conditioned on the patterns. Then, following 
Shannon's theorem, the length of the optimal code for the database
under the model is simply the negative logarithm
 of its probability. This allows us to search for
the set of patterns that best compress the database
without designing a specific coding scheme.
Our approach, which we call the \emph{Interesting Sequence Miner} (ISM)\footnote{https://github.com/mast-group/sequence-mining},
is a novel sequential pattern mining algorithm that is able to efficiently mine the most relevant sequential patterns from a database and rank them using 
an associated measure of interestingness. ISM makes use of a novel probabilistic
model of sequences, based on generating a sequence by
interleaving a group of subsequences. It is these learned component subsequences that are the patterns ISM returns. 

An approach based on 
probabilistic machine learning brings a variety of benefits,
namely, that the probabilistic model allows us to
declaratively incorporate ideas about what types
of patterns would be most useful; 
that we can
easily compose the ISM model with other types of probabilistic
models from the literature; and that we
are able to bring to bear powerful tools for inference and optimization
from probabilistic machine learning.
Inference in our model involves approximate optimization of a non-monotone submodular objective subject to a submodular coverage constraint. The necessary partition function is intractable to construct directly, however we show that it can be efficiently computed using a suitable lower bound. 
The set of sequential patterns under our model can be inferred efficiently using a \emph{structural expectation maximization} (EM) framework \cite{friedman1998bayesian}.
This is, to our knowledge, the first use of an expectation-maximization scheme for the subsequence mining problem.


On real-world datasets (\autoref{sec:numerics}), we find that ISM returns a notably more
diverse set of patterns than the 
recent  MDL methods SQS and GoKrimp (\autoref{tab:redundancy}),
while retaining similar quality.
A more diverse set of patterns is, we suggest,
especially suitable for manual examination during exploratory data analysis.
Qualitatively, the mined patterns from ISM are all
highly correlated and extremely relevant, \eg representing phrases
such as \emph{oh dear} or concepts such as \emph{reproducing kernel hilbert space}. 
More broadly, this new perspective has the potential 
to open up a wide variety of future directions for new 
modelling approaches, 
such as combining sequential pattern mining methods with hierarchical models, topic models, and nonparametric Bayesian methods.

%% file: text/related_work.tex
\section{Related Work}\label{sec:lit}

Sequential pattern mining was first introduced by \citet{agrawal1995mining} in the context of market basket analysis,
which led to a number of other algorithms for frequent subsequence, including
 GSP  
\cite{srikant1996mining}, PrefixSpan \cite{pei2001prefixspan},
 SPADE \cite{zaki2001spade}, and SPAM \cite{ayres2002sequential}.
 Frequent sequence mining suffers from \emph{pattern explosion}: a huge number of highly redundant frequent sequences are retrieved if the given minimum support threshold is too low.
One way to address this is by mining frequent closed sequences,
\ie those that have no subsequences with the same frequency,
such  as via 
the BIDE algorithm \cite{wang2004bide}.
However, even mining frequent closed sequences does not fully resolve the 
problem of pattern explosion.
We refer the interested reader to Chapter 11 of \citep{aggarwal2014frequent} for a survey of frequent sequence mining algorithms.

In an attempt to tackle this problem, modern approaches to sequence mining have used the \emph{minimum description length} (MDL) principle to find the set of sequences that best summarize the data. The GoKrimp algorithm \cite{lam2014mining} directly mines sequences that best compress a database using a MDL-based approach. The goal of GoKrimp is essentially to cover the database with as few sequences as possible, 
because the dictionary-based description length
that is used by GoKrimp favours encoding schemes that cover more long and frequent subsequences in the database.
In fact, finding the most compressing sequence in the database is strongly related to the maximum tiling problem, \ie finding the tile with largest area in a binary transaction database.

SQS-Search (SQS) \cite{tatti2012long} also uses MDL to find the set of sequences that summarize the data best: a small set of informative sequences that achieve the best compression is mined directly from the database. SQS uses an encoding scheme that explicitly punishes gaps by assigning zero cost for encoding non-gaps and higher cost for encoding larger gaps between items in a pattern. While SQS can be very effective at mining informative patterns from text, it cannot handle interleaving patterns, unlike GoKrimp and ISM, which can be a significant drawback on certain datasets \eg patterns generated by independent processes that may frequently overlap. 

In related work, \citet{mannila2000global} proposed a generative model of sequences which finds partial orders that describe the ordering relationships between items in a sequence database. Sequences are generated by selecting a subset of items from a partial order with a learned inclusion probability and arranging them into a compatible random ordering. Unlike ISM, their model does not allow gaps in the generated sequences and each sequence is only generated from a single partial order, an unrealistic assumption in practice. 

There has also been some existing research on probabilistic models for sequences, especially using Markov models. \citet{gwadera2005markov} use a variable order Markov model to identify statistically significant sequences. \citet{stolcke1993hidden} developed a structure learning algorithm for HMMs that learns both the number of states and the topology. \citet{landwehr2008modeling} extended HMMs to handle a fixed number of hidden processes whose outputs interleave to form a sequence. Wood \etal developed the sequence memoizer \cite{wood2011sequence}, a variable order Markov model with a Pitman-Yor process prior. Also, \citet{nevill1997identifying} infer a context-free grammar over sequences using the Sequitur algorithm.

%% file: text/problem_formulation.tex
\section{Mining Sequential Patterns}\label{sec:mining}
In this section we will formulate the problem of identifying a set of 
interesting sequences that are useful for explaining a sequence database.
First we will define some preliminary concepts and notation. An \emph{item} $i$ 
is an element of a universe $U = \{1,2,\dotsc,n\}$ that indexes symbols. A \emph{sequence} $S$ is simply an ordered list of items $(e_1,\dotsc,e_m)$ such that $e_i \in U \; \forall i$. A sequence $S_a = (a_1,\dotsc,a_n)$ is a subsequence of another sequence $S_b = (b_1,\dotsc,b_m)$, denoted $S_a \subset S_b$, if there exist integers $1 \le i_1 < i_2 < \dotsc < i_n \le m$ such that $a_1=b_{i_1},a_2=b_{i_2},\dotsc,a_n=b_{i_n}$ (\ie the standard definition of a subsequence). A sequence \emph{database} is merely a list of sequences $X^{(j)}$. Further, we say that a sequence $S$ is \emph{supported} by a sequence $X$ in the sequence database if $S \subset X$. Note that in the above definition each sequence only contains a single item as this is the most important and popular sequence type (\cf word sequences, protein sequences, click streams, \etc).\footnote{Note that we can easily extend our algorithm to mine sequences of sets of items (as defined in the original sequence mining paper \cite{agrawal1995mining}) by extending the subsequence operator $\subset$ to handle these more general `sequences'.} A \emph{multiset} $\mathcal{M}$ is a generalization of a set that allows elements to occur multiple times, \ie with a specific \emph{multiplicity} $\#_{\mathcal{M}}(\cdot)$. For example in the multiset $\mathcal{M} = \{a,a,b\}$, the element $a$ occurs twice and so has multiplicity $\#_{\mathcal{M}}(a) = 2$.

\subsection{Problem Formulation}
Our aim in this work is to infer a set of interesting
subsequences $\mathcal{I}$ from a database of sequences $X^{(1)},\dotsc,X^{(N)}$. Here by \emph{interesting},
we mean a set of patterns that are useful for helping a human analyst to understand the important properties of the database,
that is, interesting subsequences should reflect the most important patterns in the data, while being sufficiently concise and non-redundant
that they are suitable for manual examination.
These criteria are inherently qualitative, reflecting the fact that the goal of data mining is to build human
insight and understanding.
To quantify these criteria, 
we operationalize the notion of interesting sequence as
those sequences that best explain the underlying database under a \emph{probabilistic model} of sequences. 
 Specifically we will use a \emph{generative} model, \ie a model that starts with a set of interesting subsequences $\mathcal{I}$ and from this set generates the sequence database $X^{(1)},\dotsc,X^{(N)}$. Our goal is then to infer the most likely generating set $\mathcal{I}$ under our chosen generative model. 
 We want a model that is as simple as possible yet powerful enough to capture correlations between items in sequences. A simple such model is  as follows: iteratively sample subsequences $S$ from $\mathcal{I}$ and randomly interleave them to form the database sequence $X$. 
 If we associate each subsequence $S \in \mathcal{I}$ with a probability $\pi_S$, we can sample the indicator variable $z_S \sim \text{Bernoulli}(\pi_S)$ and include it in X if $z_S=1$. However, we may wish to include a subsequence more than once in the sequence $X$, that is, we need some way of sampling the multiplicity of $S$ in $X$. The simplest way to do this is to change our generating distribution from Bernoulli to \eg Categorical and sample the multiplicity $z_S \sim \text{Categorical}(\boldsymbol\pi_S)$ where $\boldsymbol\pi_S$ is now a vector of probabilities, with one entry for each multiplicity (up to the maximum in the database). We define the generative model formally in the next section.

\subsection{Generative Model}\label{sec:model}
As discussed in the previous section, we propose a simple directed graphical model for generating a database of
sequences $X^{(1)},\dotsc,X^{(N)}$ from a set $\mathcal{I}$ of interesting
sequences.
The generative story for our model is, independently for each sequence $X$ in the database:
\begin{enumerate}
 \item For each interesting sequence $S \in \mathcal{I}$, decide independently the number of times $S$ should be included in $X$, \ie sample the multiplicity $z_S \in \mathbb{N}_0$ as
\[ z_S \sim \text{Categorical}(\boldsymbol\pi_S), \]
where $\boldsymbol\pi_S$ is a vector of multiplicity probabilities. For clarity we present the Categorical distribution here but one could use a more general distribution if desired. 
 \item Set $\mathcal{S}$ to be the multiset with multiplicities $z_S$ of all the sequences $S$ selected for inclusion in $X$: \[\mathcal{S} \deq \{ S \,|\, z_S \ge 1 \}.\]
 \item Set $\mathcal{P}$ to be the set of all possible sequences that can be generated by interleaving together all occurrences of the sequences in the multiset $\mathcal{S}$, \ie
 \[\mathcal{P} \deq \{X  \,|\,\mathcal{S} \text{ partition of } X, S \subset X \; \forall  S \in \mathcal{S}\}.\]
 Here by \emph{interleaving} we mean the placing of items from one sequence into the gaps between items in another whilst maintaining the orders of the items imposed by each sequence.
 \item Sample $X$ uniformly from $\mathcal{P}$, \ie
 \[X \sim \mathcal{P}.\]
\end{enumerate}
Note that we never need to construct the set $\mathcal{P}$ in practice, since we only require its cardinality during inference, and we show in the next section how we can efficiently compute an approximation to $\abs{\mathcal{P}}$. We can, however, sample from $\mathcal{P}$ efficiently by merging subsequences $S \in \mathcal{S}$ into $X$ one at a time as follows: splice the elements of $S$, in order, into $X$ at randomly chosen points (here by splicing $S$ into $X$ we mean the placing of items from $S$ into the gaps between items in $X$). For example, $\mathcal{S} = \{(1,2),(3,4)\}$ will generate the set of sequences $\mathcal{P} = \{(3,4,1,2), (3,1,4,2), (3,1,2,4), (1,3,4,2),\allowbreak (1,3,2,4), (1,2,3,4)\}$.
We could of course learn a transition distribution between subsequences in our model, but we choose not to do so because we want to force the model to use $\mathcal{I}$ to explain the sequential dependencies in the data.


\subsection{Inference}\label{sec:inference}
Given a set of interesting sequences $\mathcal{I}$, let $\mathbf{z}$ denote the vector of $z_S$ for all sequences $S \in \mathcal{I}$ and similarly, let $\boldsymbol\Pi$ denote the list of $\boldsymbol\pi_S$ for all $S \in \mathcal{I}$. Assuming
$\mathbf{z}, \boldsymbol\Pi$ are fully determined, it is evident from the 
generative model that the probability of generating a database sequence $X$ is
\begin{equation*}
 p(X, \mathbf{z}|\boldsymbol\Pi) =
\begin{cases} \frac{1}{\abs{\mathcal{P}}}
\prod_{S \in \mathcal{I}}
\prod_{m=0}^{\abs{\boldsymbol\pi_S}-1} \pi_{S_m}^{[z_S=m]} \! &\text{if $X \in \mathcal{P}$,} \\
\,0 \! &\text{otherwise},
\end{cases}
\end{equation*}
where $\abs{\boldsymbol\pi_S}$ is the length of $\boldsymbol\pi_S$ and $[z_S=m]$ evaluates to $1$ if $z_S=m$, $0$ otherwise. Intuitively, it helps to think of each $\boldsymbol\pi_S$ as being an infinite vector and each $S \in \mathcal{I}$ as being augmented with a Kleene star operator, so that, for example, one can use $(1, 2)^*$ and $(3)^*$ to generate the sequence $(1, 2, 1, 3, 2)$.

Calculating the normalization constant $\abs{\mathcal{P}}$ is problematic as we have to count the number of possible distinct sequences that could be generated by interleaving together subsequences in $\mathcal{S}$. This is further complicated by the fact that $\mathcal{S}$ is a multiset and so can contain multiple occurrences of the same subsequence, which makes efficient computation of $\abs{\mathcal{P}}$ impractical. However, it turns out that we can compute a straightforward upper bound since $\abs{\mathcal{P}}$ is clearly bounded above by all possible permutations of all the items in all the subsequences $S \in \mathcal{S}$, and this bound is attained when $\mathcal{S}$ contains only distinct singleton sequences without repetition. Formally,   
\[
\abs{\mathcal{P}} \le \left(\textstyle\sum_{S \in \mathcal{S}} \abs{S}\right)! 
\]
Conveniently, this gives us a non-trivial lower-bound on the posterior $p(X, \mathbf{z}|\boldsymbol\Pi)$ which, as we will want to maximize the posterior, is precisely what we want. Moreover, the lower bound acts as an additional penalty, strongly favouring a non-redundant set of sequences (see \autoref{sec:redundancy}). 

Now assuming the parameters $\boldsymbol\Pi$ are known, we can infer 
$\mathbf{z}$ for a database sequence $X$ by maximizing the log of the lower bound  on the posterior $p(X, \mathbf{z}|\boldsymbol\Pi)$ over $\mathbf{z}$:
\begin{equation}\label{eq:logprob}
\begin{split}
&\max_{\mathbf{z}} \sum_{S \in \mathcal{I}}\sum_{m=0}^{\abs{\boldsymbol\pi_S}-1} [z_S=m] \ln(\pi_{S_m})  -\sum_{j=1}^{\sum_{S \in \mathcal{S}}\abs{S}} \ln j \\
&\st \;\; X \in \mathcal{P}. \\
\end{split}
\end{equation}
This is an NP-hard problem in general and so impractical to solve 
directly in practice. However, we will show that it can be viewed as a special case of maximizing a submodular function subject to a submodular constraint and so approximately solved using the greedy algorithm for submodular function optimization. 
Now strictly speaking the notion of a submodular function is only applicable to sets, however we will consider the  following generalization to multisets:
\begin{definition} \textit{(Submodular Multiset Function)}\label{def:submod}
Let $\Omega$ be a finite multiset and let ${\mathbb{N}_0}^{\Omega}$ denote the set of all possible multisets that are subsets of $\Omega$, 
then a function $f \colon {\mathbb{N}_0}^{\Omega} \to \mathbb{R}\,$ is \emph{submodular} if for for every $\mathcal{C}\subset\mathcal{D}\subset\Omega$ and $S \in \Omega$ with $\#_{\mathcal{C}}(S) = \#_{\mathcal{D}}(S)$ it holds that
\[
 f(\mathcal{C}\cup\{S\}) - f(\mathcal{C}) \ge f(\mathcal{D}\cup\{S\}) - f(\mathcal{D}).
\]
\end{definition}

Let us now define a function $f$ for our specific case:
let $\mathcal{T}$ be the multiset of supported interesting sequences, \ie sequences $S \in \mathcal{I}$ \st $S \subset X$ with multiplicity given by 
the maximum number of occurrences of $S$ in any partition of $X$. 
Now, define $f \colon {\mathbb{N}_0}^{\mathcal{T}} \to \mathbb{R}$ as
\[
f(\mathcal{C}) \deq \sum_{S \in \mathcal{C}}\sum_{m=0}^{\abs{\boldsymbol\pi_S}-1}[\#_{\mathcal{C}}(S)=m]\ln(\pi_{S_m}) 
-\sum_{j=1}^{\sum_{S \in \mathcal{C}}\abs{S}} \ln j
\]
and $g(\mathcal{C}) \deq \abs{\cup_{S \in \mathcal{C}} S}$. We can now re-state \eqref{eq:logprob} as: Find a non-overlapping multiset covering $\mathcal{C} \subset \mathcal{T}$ that maximizes $f(\mathcal{C})$ , \ie such that $g(\mathcal{C}) = g(\mathcal{T})$ and $f(\mathcal{C})$ is maximized. Note that $g(\mathcal{T}) = \abs{X}$ by construction. Now clearly $g$ is monotone submodular as it is a multiset coverage function, and we will show that $f$ is non-monotone submodular. To see that $f$ is submodular observe that
for $\mathcal{C}\subset\mathcal{D}, \#_{\mathcal{C}}(S) = \#_{\mathcal{D}}(S)$
\[
\begin{split}
f(\mathcal{D}\cup\{S\}) - f(\mathcal{D}) &= \ln(\pi_{S_{\#_{\mathcal{D}}(S)+1}}) -\ln(\pi_{S_{\#_{\mathcal{D}}(S)}}) \\
&\quad- \sum_{j=\sum_{D\in \mathcal{D}}\abs{D}+1}^{\sum_{D\in\mathcal{D}}\abs{D}+\abs{S}} \ln j  \\
&\le \ln(\pi_{S_{\#_{\mathcal{C}}(S)+1}}) -\ln(\pi_{S_{\#_{\mathcal{C}}(S)}}) \\
&\quad- \sum_{j=\sum_{C\in \mathcal{C}}\abs{C}+1}^{\sum_{C\in\mathcal{C}}\abs{C}+\abs{S}} \ln j  \\
&= f(\mathcal{C}\cup\{S\}) - f(\mathcal{C})
\end{split}
\]
which is precisely \autoref{def:submod}. To see that $f$ is non-monotone observe that 
\[
f(\mathcal{C}\cup\{S\}) - f(\mathcal{C}) = \ln\left(\frac{\pi_{S_{\#_{\mathcal{C}}(S)+1}}}{\pi_{S_{\#_{\mathcal{C}}(S)}}}\right)  
- \sum_{j=\sum_{C\in \mathcal{C}}\abs{C}+1}^{\sum_{C\in\mathcal{C}}\abs{C}+\abs{S}} \ln j 
\]
whose sign is indeterminate. 

Maximizing the posterior \eqref{eq:logprob} is therefore a problem of maximizing a submodular function subject to a submodular coverage constraint and can be approximately solved by applying the greedy approximation algorithm (\autoref{alg:greedy}). The greedy algorithm builds a multiset covering $\mathcal{C}$ by repeatedly 
choosing a sequence $S$ that maximizes the profit $f(\mathcal{C}\cup\{S\}) - f(\mathcal{C})$ of adding $S$ to the covering divided by the number
of items in $S$ not yet covered by the covering $g(\mathcal{C}\cup\{S\}) - g(\mathcal{C}) = \abs{S}$.
In order to minimize CPU time
spent solving the problem, we cache the
sequences and coverings for each database sequence as needed.
\begin{algorithm}
\caption{\textsc{Greedy Algorithm}}\label{alg:greedy}
\algsetup{indent=2em}
\begin{algorithmic}
\REQUIRE Database sequence $X$, supported sequences $\mathcal{T}$
\STATE Initialize multiset $\mathcal{C} \leftarrow \emptyset$
\WHILE{$g(\mathcal{C}) \neq \abs{X}$}
\STATE Choose $S \in \mathcal{T}$ maximizing $\frac{f(\mathcal{C}\cup\{S\}) - f(\mathcal{C})}{\abs{S}}$
\STATE $\mathcal{C} \leftarrow \mathcal{C}\cup \{S\}$
\ENDWHILE
\RETURN $\mathcal{C}$
\end{algorithmic}
\end{algorithm}

Note that while there are good theoretical guarantees on the approximation ratio
achieved by the greedy algorithm when maximizing a \emph{monotone} submodular set function subject to a coverage constraint (\eg $\ln\abs{X} +1$ for weighted set cover \cite{chvatal1979greedy,feige1998threshold}) the problem of maximizing a \emph{non-monotone} submodular set function subject to a coverage constraint has, to the best of our knowledge, not been studied in the literature. However, as our submodular optimization problem is an extension of the weighted set cover problem, the greedy algorithm is a natural fit and indeed we observe good performance in practice. 

\subsection{Learning}
Given a set of interesting sequences $\mathcal{I}$, consider now the case where both
variables $\mathbf{z}, \boldsymbol\Pi$ in the model are unknown. In this case
we can use the hard EM algorithm
\citep{dempster1977maximum} for parameter estimation with latent variables. The hard-EM 
algorithm in our case is merely a simple layer on top of the inference
algorithm \eqref{eq:logprob}. Suppose there are $N$ database sequences
$X^{(1)},\dotsc,X^{(N)}$
with multisets of supported interesting sequences $\mathcal{T}^{(1)},\dotsc,\mathcal{T}^{(N)}$,
then the hard EM algorithm is given in~\autoref{alg:em}
(note that $\norm{\cdot}_F$ denotes the Frobenius norm and $\pi_{S_0}$ is the probability that $S$ does not explain any sequence in the database).
To initialize $\boldsymbol\Pi$, a natural choice is simply the support (relative frequency) of each sequence.

\begin{algorithm}[tb]
\caption{\textsc{Hard-EM}}\label{alg:em}
\algsetup{indent=2em}
\begin{algorithmic}
\REQUIRE Set of sequences $\mathcal{I}$ and initial estimates
$\boldsymbol\Pi^{(0)}$
\STATE $k \leftarrow 0$
\REPEAT
\STATE $k \leftarrow k+1$
\STATE \textsc{E-step: } $\forall \, X^{(i)}$ solve \eqref{eq:logprob} to get $z_S^{(i)} \;\, \forall\, S \in
\mathcal{T}_i$
\STATE \textsc{M-step: } $\pi_{S_m}^{(k)} \leftarrow \frac{1}{N} \sum_{i=1}^N
[z_S^{(i)}=m] \quad \forall\, S \in \mathcal{I}, \; \forall\, m$ 
\UNTIL{$\norm{\boldsymbol\Pi^{(k-1)} - \boldsymbol\Pi^{(k)}}_F > \varepsilon\,$} 
\STATE Remove from $\mathcal{I}$ sequences $S$ with $\pi_{S_0} = 1$
\RETURN $\mathcal{I}, \boldsymbol\Pi^{(k)}$
\end{algorithmic}
\end{algorithm}

\subsection{Inferring new sequences}
We infer new sequences using structural EM \citep{friedman1998bayesian},  
\ie we add a candidate sequence $S'$ to $\mathcal{I}$ if doing so
improves the optimal value $\overline{p}$ of the problem \eqref{eq:logprob} averaged
across all database sequences. 
Interestingly, there are two implicit regularization
effects here. Firstly, observe from \eqref{eq:logprob} that when a new candidate $S'$ is added to the model, a corresponding term $\ln \pi_{S'_0}$ is added to the
log-likelihood of all database sequences that $S'$ does not support. For large sequence databases, this amounts to a significant penalty on candidates in practice. Secondly, observe that the last term of \eqref{eq:logprob} acts as an additional penalty, strongly favouring a non-redundant set of sequences.

To get an estimate of maximum benefit to including
candidate $S'$, we must carefully choose an initial value of
$\boldsymbol\pi_{S'}$ that is not too low, to avoid getting stuck in a local optimum.
To infer a good $\boldsymbol\pi_{S'}$, we force the candidate $S'$ to explain all
database sequences it supports by initializing $\boldsymbol\pi_{S'} = (0,1,\dotsc,1)^T$ and update $\boldsymbol\pi_{S'}$
with the probability corresponding to its actual usage once we have inferred
all the coverings. Given a set of interesting sequences $\mathcal{I}$ and corresponding probabilities $\boldsymbol\Pi$ along with database sequences $X^{(1)},\dotsc,X^{(N)}$,
each iteration of the structural EM algorithm is given in \autoref{alg:sem} below.
\begin{algorithm}
\caption{\textsc{Structural-EM} (one iteration) 
}\label{alg:sem}
\algsetup{indent=2em}
\begin{algorithmic}
\REQUIRE Sequences $\mathcal{I}$, $\boldsymbol\Pi$, optima $p^{(i)}$ of \eqref{eq:logprob} $\forall \, X^{(i)}$
\STATE Set profit  $\overline{p} \leftarrow \frac{1}{N}
\sum_{i=1}^N p^{(i)}$
\REPEAT 
\STATE Generate candidate $S'$ using \textsc{Candidate-Gen}
\STATE $\mathcal{I} \leftarrow \mathcal{I}\cup
\{S'\}, \, \boldsymbol\pi_{S'} \leftarrow (0,1,\dotsc,1)^T$ 
\STATE \textsc{E-step: } $\forall \, X^{(i)}$ solve \eqref{eq:logprob} to get $z_S^{(i)} \;\, \forall\, S \in \mathcal{T}_i$
\STATE \textsc{M-step: } $\pi_{S_m}' \leftarrow \frac{1}{N} \sum_{i=1}^N
[z_S^{(i)}=m] \quad \forall\, S \in \mathcal{I}, \; \forall\, m$
\STATE $\forall \, X^{(i)}$, solve \eqref{eq:logprob} using
$\boldsymbol\pi_S',z_S^{(i)} \;\, \forall\, S \in \mathcal{T}_i$ 
\STATE \quad to get the optimum $p^{(i)}$
\STATE Set new profit $\overline{p}' \leftarrow \frac{1}{N} \sum_{i=1}^N
p^{(i)}$
\STATE $\mathcal{I} \leftarrow \mathcal{I}\setminus \{S'\}$
\UNTIL{$\overline{p}' \le \overline{p}$} \COMMENT{until one good candidate
found} 
\STATE $\mathcal{I} \leftarrow \mathcal{I}\cup \{S'\}$
\RETURN $\mathcal{I}, \boldsymbol\Pi'$
\end{algorithmic}
\end{algorithm}

Occasionally the \textsc{Hard-EM} algorithm may assign zero probability to one or more singleton sequences and cause the greedy algorithm to not be able to fully cover a database sequence $X$ using just the interesting sequences in $\mathcal{I}$.
In this case we simply re-seed $\mathcal{I}$ with the necessary singletons. 
Finally, in practice we store the set of candidates that have been rejected by \textsc{Structural-EM} and check each potential candidate against this set for efficiency.

\subsection{Candidate generation}\label{sec:candgen}
The \textsc{Structural-EM} algorithm (\autoref{alg:sem}) requires a method to generate 
new candidate sequences $S'$ that are to be considered for inclusion in the set of interesting sequences $\mathcal{I}$. One possibility would be to use the
GSP algorithm \cite{srikant1996mining} to recursively suggest larger sequences starting from singletons,
however preliminary experiments found this was not the most efficient method. 
For this reason we take a slightly different approach and recursively
combine the interesting sequences in $\mathcal{I}$ with the \emph{highest
support first} (\autoref{alg:combine}). In this way our candidate generation
algorithm is more likely to propose viable candidate sequences earlier and in
practice we find that this heuristic works well.

\begin{algorithm}[h]
\caption{\textsc{Candidate-Gen}}\label{alg:combine}
\algsetup{indent=2em}
\begin{algorithmic}
\REQUIRE Sequences $\mathcal{I}$, cached supports $\boldsymbol\sigma$, queue length $q$
\IF{$\nexists$ priority queue $\mathcal{Q}$ for $\mathcal{I}$}
\STATE Initialize $\boldsymbol\sigma$-ordered priority queue $\mathcal{Q}$
\STATE Sort $\mathcal{I}$ by decreasing sequence support using $\boldsymbol\sigma$
\FOR{all ordered pairs $S_1,S_2 \in \mathcal{I}$, highest
ranked first}
\STATE Generate candidate $S' = S_1 S_2$
\STATE Cache support of $S'$ in $\boldsymbol\sigma$ and add $S'$ to $\mathcal{Q}$
\STATE \textbf{if} $\abs{\mathcal{Q}}=q$ \textbf{break} 
\ENDFOR
\ENDIF
\STATE Pull highest-ranked candidate $S'$ from $\mathcal{Q}$
\RETURN $S'$
\end{algorithmic}
\end{algorithm}

\subsection{Mining Interesting Sequences}\label{sec:ism}
Our complete interesting sequence mining (ISM) algorithm is given in \autoref{alg:ism}.
\begin{algorithm}[h]
\caption{\textsc{ISM} (Interesting Sequence Miner)}\label{alg:ism}
\algsetup{indent=2em}
\begin{algorithmic}
\REQUIRE Database of sequences $X^{(1)},\dotsc,X^{(N)}$
\STATE Initialize $\mathcal{I}$ with singletons, $\boldsymbol\Pi$ with their
supports
\WHILE{not converged}
\STATE Add sequences to $\mathcal{I}, \boldsymbol\Pi$ using \textsc{Structural-EM}
\STATE Optimize parameters for $\mathcal{I}, \boldsymbol\Pi$ using \textsc{Hard-EM}
\ENDWHILE
\RETURN $\mathcal{I},\boldsymbol\Pi$
\end{algorithmic}
\end{algorithm}
Note that the \textsc{Hard-EM} parameter optimization step need not be
performed at every iteration, in fact it is more efficient to
suggest several candidate sequences before optimizing the parameters. As all operations on database sequences in our algorithm are trivially parallelizable, we perform the $E$ and $M$-steps in both
the hard and structural EM algorithms in parallel.
 
\subsection{Interestingness Measure}\label{sec:interestingness}
Now that we have inferred the model variables $\mathbf{z},\boldsymbol\Pi$, we are able to use them to rank the retrieved sequences in $\mathcal{I}$. There are two natural rankings one can employ, and both have their strengths and weaknesses. The obvious approach is to rank each sequence $S \in \mathcal{I}$ according to its probability under the model $\boldsymbol\pi_S$, however this has the disadvantage of strongly favouring frequent sequences over rare ones, an issue we would like to avoid. 
An alternative is to rank the retrieved sequences according to their \emph{interestingness} under the model, that is the ratio of database sequences they explain to database sequences they support. One can think of interestingness as a measure of how necessary the sequence is to the model: the higher the interestingness, the more supported database sequences the sequence explains. Thus interestingness provides a more balanced measure than probability, at the expense of missing some frequent sequences that only explain some of the database sequences they support. We define interestingness formally as follows. 
\begin{definition}
The \emph{interestingness} of a sequence $S\in \mathcal{I}$ retrieved by ISM
(\autoref{alg:ism}) is defined as
\[
 int(S) = \frac{\sum_{i=1}^N
[z_S^{(i)}\ge 1]}{supp(S)}
\]
and ranges from $0$ (least interesting) to $1$ (most interesting).
\end{definition}
Any ties in the ranking can be broken using the sequence probability $p(S \subset X) = p(z_S \ge 1) = 1 - \pi_{S_0}$.

\subsection{Correspondence to Existing Models}

There is a close and well-known connection between probabilistic modelling and the minimum description length
principle used by SQS and GoKrimp
 (see \citet{mackay:book}, \S 28.3 for a particularly nice explanation).
Given a probabilistic model $p(X | \PiB, \calI)$ of a single 
database sequence $X$, by Shannon's theorem the optimal code
for the model
will encode $X$ using approximately $-\log_2 p(X | \PiB, \calI)$ bits. So by finding a set of patterns
that maximizes the probability of the data, we are also finding patterns that minimize description length.
Conversely, any encoding scheme implicitly defines a probabilistic model. Given an encoding scheme $E$ that assigns each database sequence $X$ to a string of $L(X)$ bits, 
we can define $p(X | E) \propto 2^{-L(X)},$ and then $E$ is an optimal code for $p(X|E)$. Interpreting
the previous subsequence mining methods in terms of their implicit probabilistic models provides interesting
insights into these methods.

The encoding of a database sequence used by 
SQS can be interpreted as a probabilistic model $p(X, \zB | \PiB, \calI),$ where
the SQS analog of  $p(X,\zB | \PiB,\calI)$ is similar to \eqref{eq:logprob}
with 
\[
 \pi_{S_m} = \left( \frac{ \sum_{i=1}^N z_S^{(i)} }{ \sum_{I \in \mathcal{I}} \sum_{j=1}^N z_I^{(j)} } \right)^m,
 \]
along with additional terms that correspond to the
description lengths for indicating the presence and absence of gaps in the usage of a sequence $S$.
 Additionally, SQS contains
an explicit penalty for the encoding of the set of patterns $\calI$, that encourages a smaller number of patterns. In a probabilistic model, this can be interpreted as a prior distribution 
$p(\calI)$ over patterns. There is also a prior distribution on the content of the
patterns, similar to a unigram model, which 
encourages the patterns to contain more common elements.


Similarly, GoKrimp uses a variant of the above model,
where instead we have 
\[
\pi_{S_m} = \left( \frac{ \sum_{i=1}^N z_S^{(i)} + \abs{\{T \in \mathcal{I} \,|\, S \subset T \}} }{ \sum_{I \in \mathcal{I}} \sum_{j=1}^N z_I^{(j)} + \abs{\{T \in \mathcal{I} \,|\, I \subset T \}} } \right)^m.
\]
In addition, the description length used by GoKrimp also has a gap cost that penalizes sequences with large gaps. 
GoKrimp employs a greedy heuristic to find the most compressing sequence: an empty sequence $S$ is iteratively extended by the most frequent item that is statistically dependent on $S$.
 ISM, by contrast, iteratively extends sequences by the most frequent sequence in its candidate generation step which enables it to quickly generate large candidate sequences (\autoref{sec:candgen}).
 We did consider performing a statistical test between a sequence and its extending sequence, however this proved computationally prohibitive.

The differences between these models and ISM are:
\begin{itemize}
\item \emph{Interleaving.}  SQS cannot mine subsequences 
that are interleaved and thus struggles on datasets which consist mainly of interleaved subsequences 
(for illustration, see \autoref{sec:interpretability}).
GoKrimp handles interleaving using a pointer scheme that explicitly encodes the location of the subsequence within the database.
In ISM, the partition function $|\mathcal{P}|$ allows us to handle interleaving of subsequences without needing
to explicitly encode positions, and also serves as an additional penalty on the number of elements in the subsequences used
to explain a database sequence.
\item \emph{Gap penalties.} Both SQS and GoKrimp explicitly punish gaps in sequential patterns.
Adding such a penalty would require only a trivial modification to the algorithm, namely, updating the cost function in \autoref{alg:greedy}.
We did not pursue this as we observe excellent results without it (\autoref{sec:numerics}). 
\item \emph{Encoding the set of patterns.}
Both SQS and GoKrimp contain an explicit penalty term
for the description length of the pattern database,
which corresponds to a prior distribution $p(\calI)$ over patterns.
In our experiments with ISM, we did not find in practice
that an explicit prior distribution $p(\calI)$ was necessary
for good results. It would be possible to incorporate it
with a trivial change to the ISM algorithm, in particular, 
when computing the score improvement of a new candidate
in the structural EM step.
\item \emph{Encoding pattern absence.}
Also, observe that, if we view ISM as an MDL-type method, 
not only the presence of a pattern, but also the absence of it is explicitly encoded (in the form of $\pi_{S_0}$ in \eqref{eq:logprob}). As a result, there is an implicit penalty for adding too many patterns to the model and one does not need to use a code table which would serve as an explicit penalty for greater model complexity.
\end{itemize}

%% file: text/experiments.tex
\section{Numerical Experiments}\label{sec:numerics}
In this section we perform a comprehensive quantitative and qualitative
evaluation of ISM. On synthetic datasets we show that ISM returns a list of sequential patterns that 
is largely non-redundant, contains few spurious correlations and scales linearly with the number of sequences in the dataset. On a set of real-world datasets we show that ISM finds patterns that are consistent, interpretable and highly relevant to the problem at hand. Moreover, we show that ISM is able to mine patterns that achieve good accuracy when used as binary features for real-world classification tasks.

\begin{table}[b]
 \small\centering
  \begin{tabular}{lrr|rr}
  \toprule
  Dataset & Uniq.\ Items & Sequences & Subseq.\dag & Runtime \\
  \midrule
  Alice & $2,619$ & $1,638$ & $123$ & $114$ min \\
  Gazelle & $497$ & $59,601$ & $727$ & $582$ min \\
  JMLR & $3,846$ & $788$ & $361$ & $230$ min \\
  Sign & $267$ & $730$ & $159$ & $31$ min \\
  aslbu & $250$ & $424$ & $144$ & $4$ min \\
  aslgt & $94$ & $3,464$ & $57$ & $19$ min \\
  auslan2 & $16$ & $200$ & $10$ & \textgreater$1$ min \\
  context & $94$ & $240$ & $19$ & $7$ min \\
  pioneer & $178$ & $160$ & $86$ & $3$ min \\
  skating & $82$ & $530$ & $70$ & $9$ min \\
    \bottomrule
  \end{tabular}
\caption{Summary of the real datasets used and ISM results after $1,000$
iterations. \dag\,excluding singleton subsequences.}
\label{tab:datasets}
\end{table}

\boldpara{Datasets}
We use ten real-world datasets in our numerical evaluation (see \autoref{tab:datasets}). The Alice dataset consists of the text of Lewis Carrol's Alice in Wonderland, tokenized into $1,638$ sentences using the Stanford Document Preprocessor \cite{manning2014stanford} with stop words and punctuation deliberately retained. The Gazelle dataset consists of $59,601$ sequences of clickstream data from an e-commerce website used in the KDD-CUP 2000 competition \cite{kohavi2000kdd}. The JMLR dataset consists of $788$ abstracts from the Journal of Machine Learning Research and has previously been used in the evaluation of the SQS and GoKrimp algorithms \cite{lam2014mining,tatti2012long}. Each sequence is a list of stemmed words from the text with stop words removed. The Sign dataset is a list of $730$ American sign language utterances where each utterance contains a number of gestural and grammatical fields \cite{papapetrou2005discovering}. The last six datasets listed in \autoref{tab:datasets} were first introduced in \cite{moerchen2010robust} to evaluate classification accuracy when mined sequential patterns are used as features. The datasets were converted from time interval sequences into sequences of items by considering the start and end of each unique interval as distinct items and ordering the items according to time.

\begin{figure}[tb]
  \centering
  \includegraphics[scale=0.35]{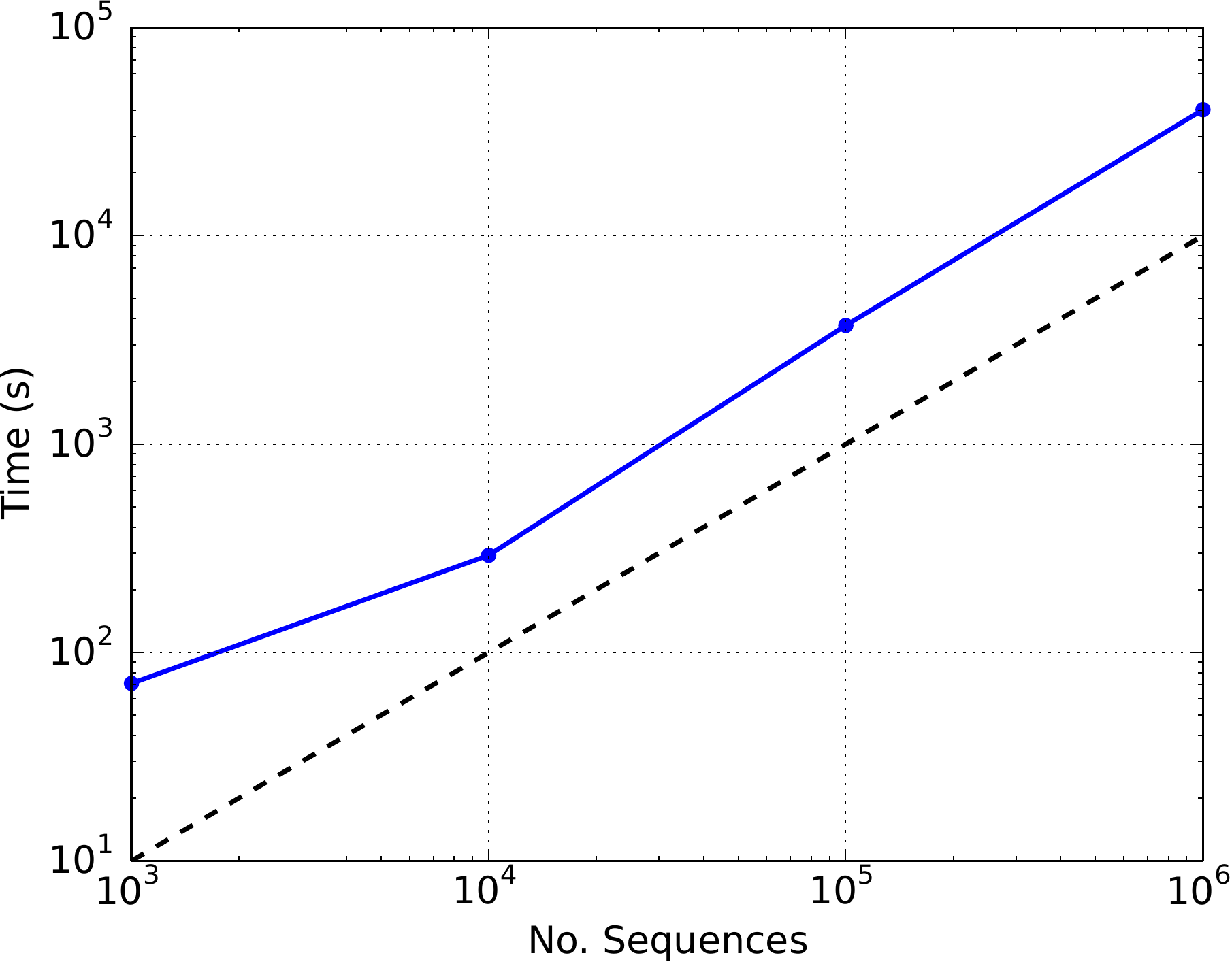}%
  \caption{ISM scaling as the number of sequences in our synthetic database
    increases.}%
  \label{fig:scaling}%
\end{figure}

\boldpara{ISM Results}
We ran ISM on each dataset
for $1,000$ iterations with a priority queue size of $100,000$ candidates. The runtime and number of non-singleton sequential patterns returned by ISM is given in the right-hand side of \autoref{tab:datasets}. 
We also investigated the scaling of ISM as the number
of sequences in the database increases, using the model trained on the
Sign dataset from \autoref{sec:spur} to generate synthetic sequence
databases of various sizes. We ran ISM for $100$ iterations on these
databases and one can see in \autoref{fig:scaling} that the scaling is
linear as expected. All experiments were performed on a machine with $16$ Intel Xeon E5-2650 $2.60$Ghz CPUs and 128GB of RAM. 

\boldpara{Evaluation criteria}
We will evaluate ISM along with SQS, GoKrimp and BIDE according to the following criteria:
\begin{enumerate}
\item \textit{Spuriousness} -- to assess the degree of spurious correlation in the mined set of sequential patterns.
\item \textit{Redundancy} -- to measure how redundant the mined set of patterns is.
\item \textit{Classification Accuracy} 
-- to measure the usefulness of the mined patterns.
\item \textit{Interpretability} 
-- to informally assess how meaningful and relevant the mined patterns actually are.
\end{enumerate}

\begin{figure}
 \centering
 \includegraphics[scale=0.35]{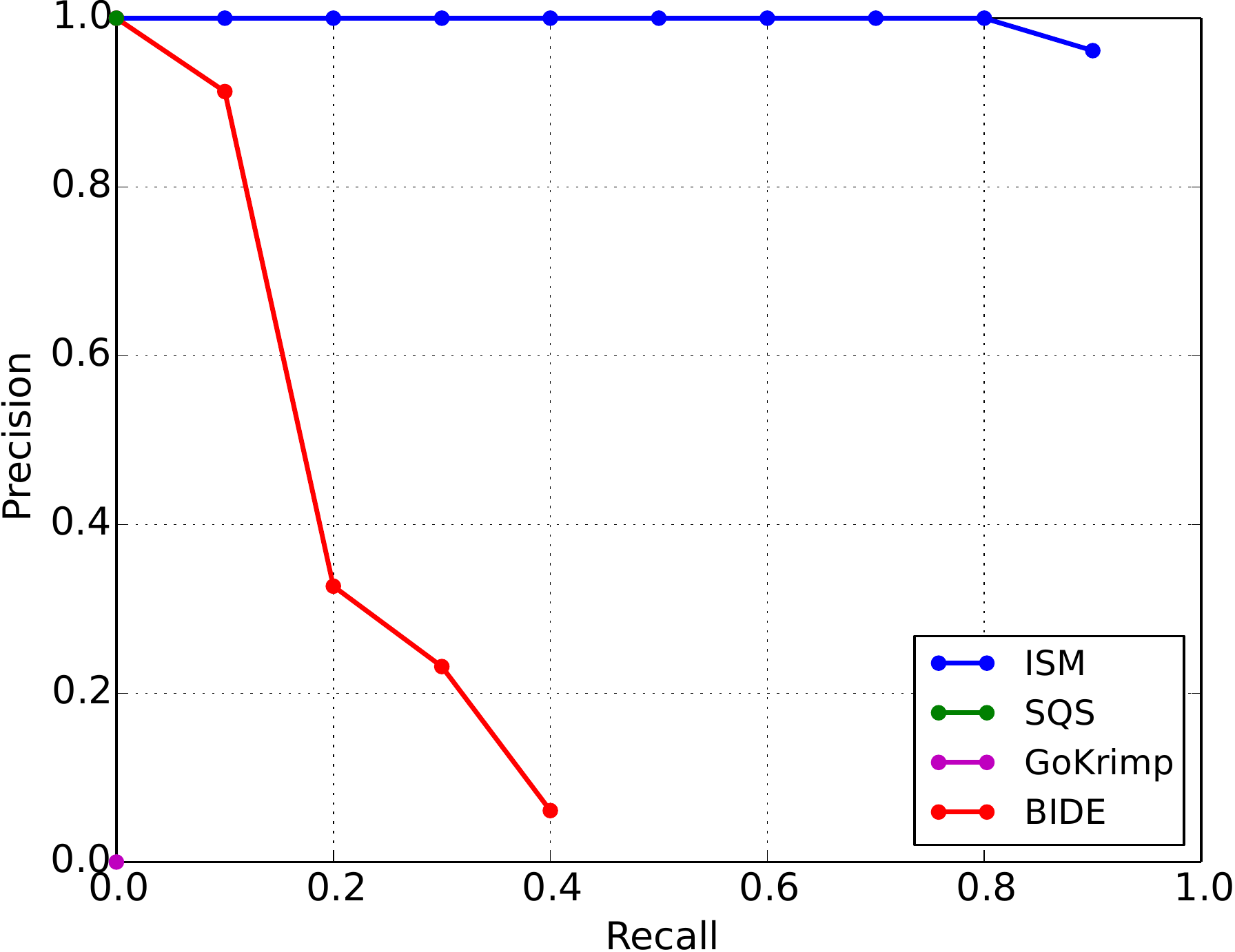}
  \caption{Precision against recall for
each algorithm on our synthetic database, using
the top-$k$ patterns as a threshold. 
Note that SQS is a single point at the top-left and GoKrimp has near zero precision and recall. 
Each plotted curve is the 11-point
interpolated precision\protect\footnotemark[3].}
\label{fig:background}
\end{figure}

\footnotetext[3]{\ie the interpolated precision at 11 equally spaced recall points between $0$ and $1$ (inclusive), see \cite{manning2008introduction}, \S8.4 for details.}

\subsection{Pattern Spuriousness}\label{sec:spur}
The sequence-cover formulation of the ISM algorithm \eqref{eq:logprob} naturally favours adding sequences to the model whose items co-occur in the sequence database. One would therefore expect ISM to largely avoid suggesting sequences of uncorrelated items and so return more meaningful patterns. To verify
this is the case and validate our inference procedure, we check if ISM is able to recover the sequences it used to generate a synthetic database. To obtain a realistic synthetic database, we sampled $10,000$ sequences from the ISM generative model trained on the Sign dataset (\cf \autoref{sec:model}). We were then able to measure the precision and recall for each algorithm, \ie the fraction of mined patterns that are generating and the fraction of generating patterns that are mined, respectively. 
\autoref{fig:background} shows the
precision-recall curve for ISM, SQS, GoKrimp and BIDE using the top-$k$ mined sequences (according to each algorithms ranking) as a threshold. One can clearly see that ISM was able to mine almost all the generating patterns and almost all the patterns mined were generating, despite the fact that the generated database will contain many subsequences not present in the original dataset due to the nature of our `subsequence interleaving' generative model. This not only provides a good validation of ISM's inference procedure and underlying generative model but also demonstrates that ISM returns few spurious patterns.
For comparison, SQS returned a very small set of generating patterns and GoKrimp returned many patterns that were not generating. The set of top-$k$ patterns mined by BIDE contained successively less generating patterns as $k$ increased. It is not our intention to draw conclusions about the performance of the other algorithms as this experimental setup naturally favours ISM.   
Instead, we compare the patterns from ISM with those from SQS and GoKrimp on real-world data in the next sections.

\begin{table*}[tb]
 \small\centering
  \begin{tabular}{lrrrrrrrrrrrrrrr}
  \toprule &
\multicolumn{3}{c}{Alice} & \multicolumn{3}{c}{Gazelle} & \multicolumn{3}{c}{JMLR} & \multicolumn{3}{c}{Sign} & \multicolumn{3}{c}{aslbu} \\
\cmidrule(r){2-4} \cmidrule{5-7} \cmidrule(l){8-10} \cmidrule(l){11-13} \cmidrule(l){14-16}
 & ISD & CS & Items & ISD & CS & Items & ISD & CS & Items & ISD & CS & Items & ISD & CS & Items \\
\cmidrule(r){2-4} \cmidrule{5-7} \cmidrule(l){8-10} \cmidrule(l){11-13} \cmidrule(l){14-16}
  \textbf{ISM} & $\mathbf{2.00}$ & $\mathbf{0.00}$ & $\mathbf{94}$ & $3.36$ & $\mathbf{0.00}$ & $167$ & $\mathbf{1.84}$ & $\mathbf{0.00}$ & $\mathbf{96}$ & $\mathbf{3.64}$ & $\mathbf{0.00}$ & $\mathbf{113}$ & $\mathbf{2.24}$ & $\mathbf{0.00}$ & $\mathbf{110}$ \\
  SQS & $1.76$ & $0.10$ & $72$ & $\mathbf{4.24}$ & $0.38$ & $\mathbf{183}$ & $1.82$ & $0.02$ & $92$ & $1.26$ & $0.94$ & $57$ & *$1.89$ & *$0.11$ & *$61$ \\
  GoKrimp & $1.24$ & $0.10$ & $52$ & *$4.51$ & *$0.05$ & *$176$ & *$1.40$ & *$0.10$ & *$30$ & $1.72$ & $0.24$ & $63$ & *$2.00$ & *$0.00$ & *$18$ \\
  BIDE & $1.00$ & $0.36$ & $29$ & $1.00$ & $0.36$ & $26$ & $1.00$ & $0.18$ & $12$ & $1.00$ & $0.60$ & $15$ & $1.00$ & $0.00$ & $26$ \\
  \midrule &
  \multicolumn{3}{c}{aslgt} & \multicolumn{3}{c}{auslan2} & \multicolumn{3}{c}{context} & \multicolumn{3}{c}{pioneer} & \multicolumn{3}{c}{skating} \\
\cmidrule(r){2-4} \cmidrule{5-7} \cmidrule(l){8-10} \cmidrule(l){11-13} \cmidrule(l){14-16}
 & ISD & CS & Items & ISD & CS & Items & ISD & CS & Items & ISD & CS & Items & ISD & CS & Items \\
\cmidrule(r){2-4} \cmidrule{5-7} \cmidrule(l){8-10} \cmidrule(l){11-13} \cmidrule(l){14-16}
  \textbf{ISM} & $\mathbf{2.08}$ & $0.20$ & $\mathbf{94}$ & *$\mathbf{2.40}$ & *$1.0$ & *$\mathbf{14}$ & *$\mathbf{2.16}$ & *$\mathbf{0.47}$ & *$35$ & $\mathbf{2.04}$ & $\mathbf{0.00}$ & $\mathbf{102}$ & $\mathbf{2.12}$ & $0.72$ & $\mathbf{73}$ \\
  SQS & $1.96$ & $0.28$ & $86$ & *$1.42$ & *$1.17$ & *$12$ & $2.14$ & $0.90$ & $\mathbf{64}$ & $1.64$ & $0.40$ & $78$ & $1.62$ & $0.84$ & $46$ \\
  GoKrimp & *$2.00$ & *$0.00$ & *$89$ & *$2.00$ & *$\mathbf{0.25}$ & *$8$ & *$2.07$ & *$0.52$ & *$51$ & *$1.82$ & *$0.00$ & *$33$ & *$1.90$ & *$\mathbf{0.29}$ & *$64$ \\
  BIDE & $1.00$ & $0.00$ & $22$ & *$1.00$ & *$3.16$ & *$6$ & $1.00$ & $1.72$ & $12$ & $1.02$ & $0.06$ & $32$ & $1.00$ & $1.06$ & $17$ \\
  \bottomrule
  \end{tabular}
\caption{Average inter-sequence distance (ISD), average no.\ containing sequences (CS) and no.\ unique items for the top $50$ non-singleton sequences returned by the algorithms from the datasets. Larger inter-sequence distances and smaller no.\ containing sequences indicate less redundancy. *\,returned less than $50$ non-singleton sequences.}
\label{tab:redundancy}
\end{table*}

\subsection{Pattern Redundancy}\label{sec:redundancy} 
We now turn our attention to evaluating how redundant the sets of sequential patterns returned by ISM, SQS, GoKrimp and BIDE actually are. A suitable measure of redundancy for a single sequence is the minimum edit distance between it and the other mined sequences in the set. Averaging this distance across all sequences in the set, we obtain the \emph{average inter-sequence distance} (ISD). Similarly, we can also calculate the average number of sequences containing other mined sequences in the set (CS), which provides us with another measure of redundancy. Finally, we can also look at the number of unique items present in the set of mined sequences which gives us an indication of how diverse it is. We ran ISM, SQS, GoKrimp and BIDE on all the datasets in \autoref{tab:datasets} and report the results of the three aforementioned redundancy metrics on the top $50$ non-singleton sequential patterns for each algorithm in \autoref{tab:redundancy}. One can see that on average the top ISM sequences have a larger inter-sequence distance, smaller number of containing sequences and larger number of unique items, clearly demonstrating they are less redundant than SQS, GoKrimp and BIDE. Predictably, the top BIDE sequences are the most redundant, with an average inter-sequence distance of $1.00$. 

\begin{figure*}
 \centering
 \includegraphics[scale=0.35]{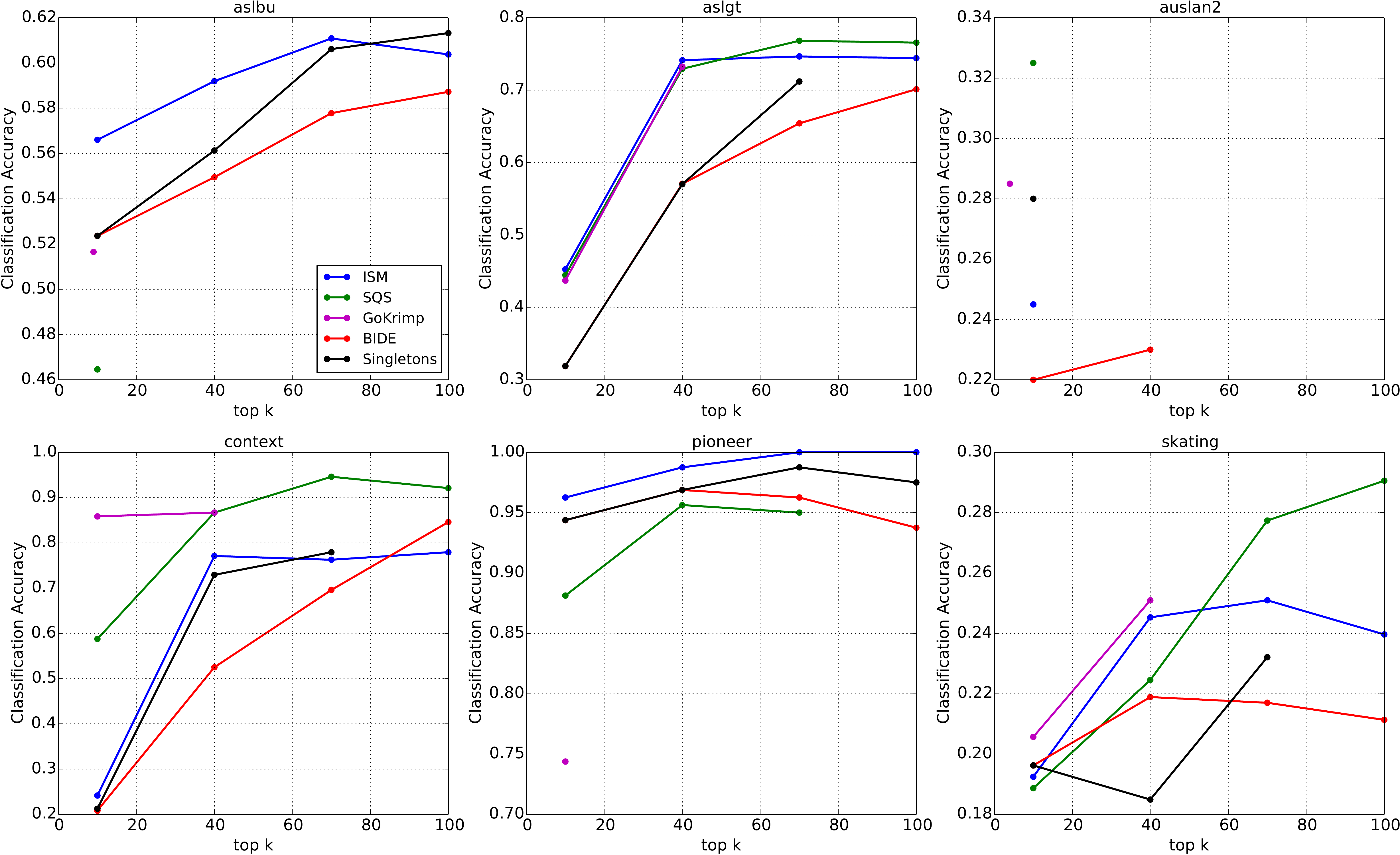}
  \caption{Linear SVM classification accuracy using the top-$k$ sequences returned by each algorithm as binary features. ISM shows consistently good performance, comparable to SQS and GoKrimp.}
\label{fig:classification}
\end{figure*}

\subsection{Classification Accuracy}
A key property of any set of patterns mined from data is its usefulness in real-world applications. To this end, in keeping with previous work \cite{lam2014mining}, we will focus on classification tasks as they are some of most important applications of pattern mining algorithms. Specifically we will consider the task of classifying sequences in a database using mined sequential patterns as binary features. We therefore performed $10$-fold cross validation using a Support Vector Machine (SVM) classifier on the six classification datasets from \autoref{tab:datasets} with the top-$k$ patterns mined by ISM, SQS, GoKrimp and BIDE as features. We used the linear classifier from the libSVM library \cite{libSVM} with default parameters. Additionally, we used the top-$k$ most frequent singleton patterns as a baseline for the classification tasks. The resulting plots of $k$ against classification accuracy for all the datasets and algorithms are given in \autoref{fig:classification}. One can see that the patterns mined by SQS perform best, exhibiting the highest classification accuracy on four out of the six datasets, closely followed by ISM and GoKrimp, which performs surprisingly well considering it struggles to return more than $50$ patterns. All three consistently outperform BIDE and the singletons baseline which exhibit similar performance to each other. We therefore conclude that the sequential patterns mined by ISM can indeed be useful in real-world applications.     

\begin{table*}
  \begin{tabular}{cccc}
  \toprule
  ISM & SQS & GoKrimp & BIDE \\
  \cmidrule(r){1-1} \cmidrule{2-2} \cmidrule(l){3-3} \cmidrule(l){4-4}
  support vector machin & support vector machin & support vector machin & algorithm algorithm \\
  real world & machin learn & machin learn & learn learn \\
  larg scale & state art & real world & learn algorithm \\
  high dimension & data set & state art & algorithm learn \\
  state art & bayesian network & high dimension & data data \\
  first second & larg scale & reproduc hilbert space & learn data  \\
  reproduc kernel hilbert space & nearest neighbor & experiment result & model model \\
  maximum likelihood & decis tree & supervis learn & problem problem \\
  wide rang & neural network & neural network & learn result \\
  gene express & cross valid & compon analysi & problem algorithm \\
  princip compon analysi & featur select & well known & method method \\
  random field & graphic model & support vector & algorithm result  \\
  maximum entropi & real world & base result & data set \\
  low dimension & high dimension & paper investig & learn learn learn \\
  blind separ & mutual inform & data demonstr & learn problem \\
  wide varieti & sampl size & hilbert space & learn method \\
  acycl graph &  learn algorithm & such paper & algorithm data \\ 
  turn out & princip compon analysi & algorithm demonstr & learn set \\
  markov chain & logist regress & learn result & problem learn \\
  leav out & model select & learn experi & algorithm algorithm algorithm \\
  \bottomrule
  \end{tabular}
   \caption{The top twenty non-singleton sequences as found by ISM, SQS, GoKrimp and BIDE for the JMLR dataset.}
    \label{tab:jmlr}
\end{table*}

\subsection{Pattern Interpretability}\label{sec:interpretability} 
For the two text datasets in \autoref{tab:datasets} we can directly interpret the mined patterns and informally assess how meaningful and relevant they are.

\boldpara{JMLR Dataset} We compare the top-$20$ non-singleton patterns mined by ISM, SQS, GoKrimp and BIDE in \autoref{tab:jmlr}. It is immediately obvious from the table that the BIDE patterns are almost exclusively permutations of frequent items and so uninformative. For this reason we omit BIDE from consideration on the next dataset. The patterns mined by ISM, SQS and GoKrimp are all very informative, containing technical concepts such as \emph{support vector machine} and commonly used phrases such as \emph{state (of the) art}.

\begin{table*}
  \begin{tabular}{ccc|ccc}
  \toprule
  ISM & SQS & GoKrimp & Exclusive ISM & Exclusive SQS & Exc.\ GoKrimp \\
  \cmidrule(r){1-1} \cmidrule{2-2} \cmidrule(l){3-3} \cmidrule(l){4-4} \cmidrule(l){5-5} \cmidrule(l){6-6} 
  she herself & ! ' & ` ' & she herself & ? ' & ` ' \\
  mock turtle & , and & , and & ( ) & the mock turtle & , said . \\
  ( ) & ? ' & , said . & he his & the march hare & said the . \\
  went on & . ' & mock turtle & as spoke & * * * * & of the .\\
  `` '' & the mock turtle & said the . & had back & , ' said alice & i n't ' \\
  ca n't & the march hare & in a & just when & , you know & march hare \\
  he his & * * * * & of the . & off head & it was & alice .\\
  looked at & , ' said alice & i n't ' & she at once & the white rabbit & what ? \\
  had been & the queen & i 'm & oh dear & , ' & you know \\
  must be & , you know & march hare & never before & ; and & ` ! \\
  at last & went on & went on & join dance & she had & you ? \\
  as spoke & it was & a little & might well & beau -- ootiful soo -- oop ! & , , ,\\
  looking at & the white rabbit & ! ' & if 'd & i 've & oh , ! \\
  had back & , ' & alice . & such thing & minute or two & i ' \\
  just when & ; and & to herself & 've seen & there was & alice ; \\ 
  off head & a little & as she & do n't know what & ` well , & , ! \\
  she at once & i 'm & what ? & going into & -- ' & the . \\
  oh dear & do n't & you know & too much & in a tone & alice , \\ 
  more than & she had & the queen & soon found & soo -- oop of the e -- e -- & it : \\
  never before & beau -- ootiful soo -- oop ! & the hatter & took its & , ' said the king & and she \\
  \bottomrule
  \end{tabular}
   \caption{The top twenty non-singleton sequences as found by ISM, SQS and GoKrimp for the Alice dataset as well as those found by ISM but not SQS/GoKrimp, SQS not ISM/GoKrimp and GoKrimp not ISM/SQS.}
    \label{tab:alice}
\end{table*}

\boldpara{Alice Dataset}
We compare the top twenty-$20$ non-singleton patterns mined by ISM, SQS and GoKrimp in the first three columns \autoref{tab:alice}. This time, one can clearly see that the patterns mined by ISM are considerably more informative. They contain collocated words and phrases such as \emph{mock turtle} and \emph{oh dear}, correlated words such as \emph{as spoke} and \emph{off head}, as well as correlated punctuation such as \emph{( )} and \emph{`` ''}. Both SQS and GoKrimp on the other hand mine collocated words with spurious punctuation and stop words, \eg prepending \emph{the} to nouns and commas to phrases. To further illustrate this notable difference, we also show the top-$20$ non-singleton patterns that are exclusive to each algorithm (\ie found by ISM but not SQS/GoKrimp, \etc) in the last three columns of \autoref{tab:alice}. One can clearly see that GoKrimp has the least informative exclusive patterns, predominantly combinations of stop words and punctuation, SQS mostly prepends and appends informative exclusive patterns with punctuation and stop words, whereas ISM is the only algorithm that just returns purely correlated words. Note that SQS in particular struggles to return patterns such as balanced parentheses, since it punishes the large gaps between them and cannot handle interleaving them with the patterns they enclose. Here we can really see the power of the statistical model underlying ISM as it is able to discern spurious punctuation from genuine phrases.

\begin{figure}
 \centering
 \includegraphics[scale=0.35]{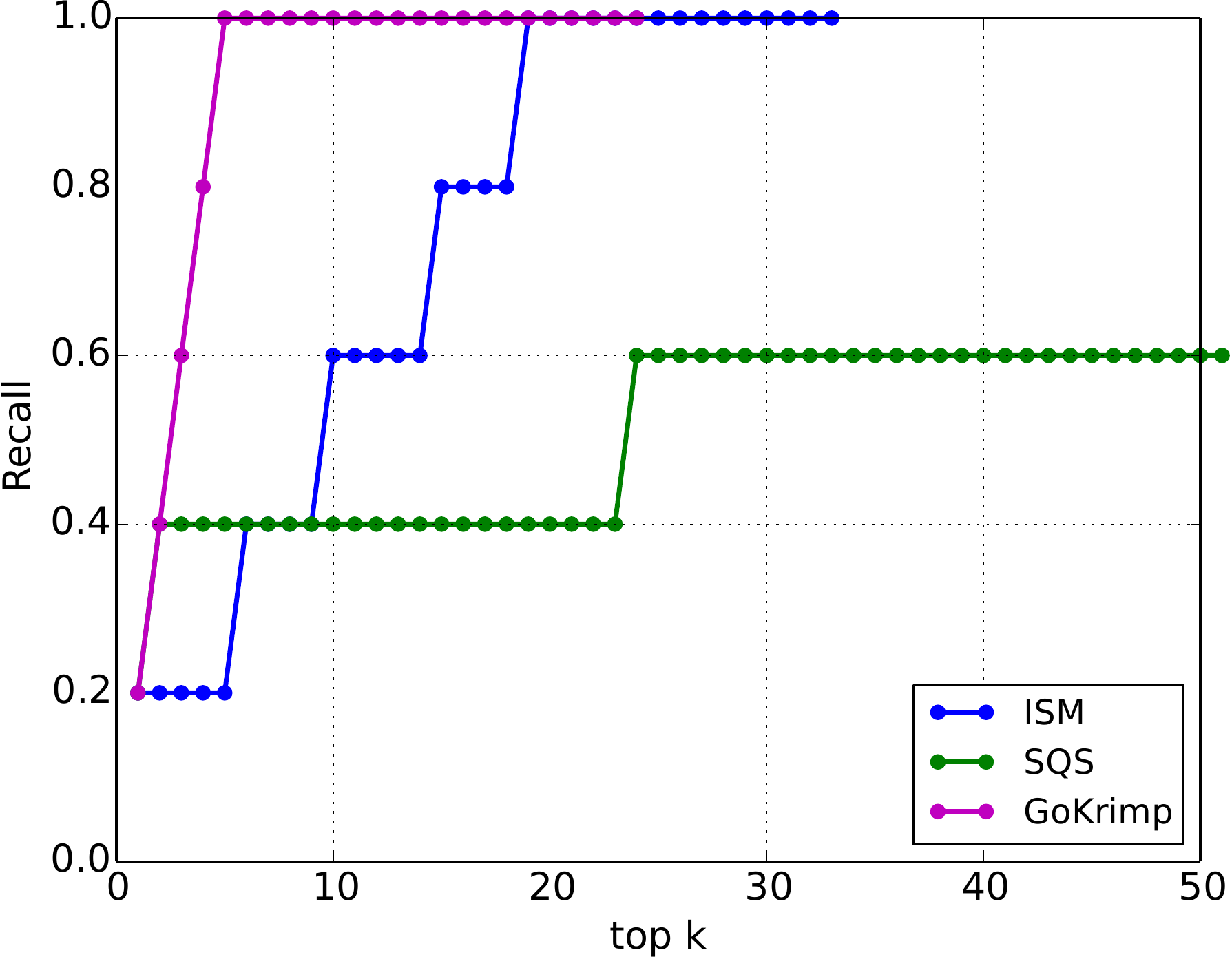}
  \caption{Recall for
each algorithm on the synthetic parallel dataset, using
the top-$k$ (first-$k$ for SQS) patterns as a threshold. Note that SQS maintains a recall level of $0.6$ for the remaining patterns (up to $k=403$, not shown for clarity).}
\label{fig:parallel}
\end{figure}

\boldpara{Parallel Dataset}
Finally, we consider a synthetic dataset that demonstrates the ability of ISM to handle interleaving patterns. Following \cite{lam2014mining}, we generate a synthetic dataset where each item in the sequence is generated by five independent parallel processes, \ie each process $i$ generates one item from a set of five possible items $\{a_i,b_i,c_i,d_i,e_i\}$ in order. In each step, the generator chooses $i$ at random and generates an item using process $i$, until the sequence has length $1,000,000$. The sequence is then split into $10,000$ sequences of length $100$. For this dataset we know that all mined sequences containing a mixture of items from different processes are spurious. This enables us to calculate recall, \ie the fraction of processes present in the set of true patterns mined by each algorithm. We plot the recall for the top-$k$ patterns mined by ISM and GoKrimp in \autoref{fig:parallel} and the first-$k$ patterns mined by SQS (as it was still running after seven days). One can see that while ISM and GoKrimp are able to mine true patterns from all processes, SQS only returns patterns from $3$ of the $5$ processes.

%% file: text/conclusions.tex
\section{Conclusions}

In this paper, we have taken a probabilistic machine learning
approach to the subsequence mining problem.
We presented a novel subsequence interleaving model,
called the Interesting Sequence Miner, that infers subsequences which best compress a sequence database without having to design a MDL encoding scheme. We demonstrated the efficacy of our approach on both synthetic and real-world datasets,
showing that ISM returns a more diverse set of patterns
than previous approaches while retaining comparable quality.
In the future we would like to extend our approach to the many promising application areas as well as considering 
more advanced techniques for parallelization.